\documentclass[runningheads]{llncs}

\usepackage{eccv}

\usepackage{eccvabbrv}

\usepackage{graphicx}
\usepackage{booktabs}
\usepackage{multirow}
\usepackage{amsmath}
\usepackage{amssymb}

\usepackage{placeins}
\usepackage[accsupp]{axessibility}  % Improves PDF readability for those with disabilities.

\usepackage{hyperref}

\usepackage{orcidlink}

\newif\ifincludeappendix
\includeappendixtrue

\begin{document}

% ---------------------------------------------------------------
\title{Task-Specific Feature Fusion Method for Multi-Task Affective Behavior Analysis}
\titlerunning{Task-Specific Feature Fusion}

% TODO FINAL: Replace with the final author list and email.
\author{
	Jiajun Sun
	\and
	Zhe Gao\textsuperscript{\dag}
}

\authorrunning{J. Sun and Z. Gao}

\institute{
	Shanghai Normal University, Shanghai, China\\
	\email{\{ora942878@gmail.com, zgao0911@shnu.edu.cn\}}
}

\maketitle

\begin{abstract}
	
	The 11th Affective Behavior Analysis in-the-wild (ABAW11) Multi-Task Learning Challenge requires a unified system to predict valence-arousal, categorical expressions, and facial action units from the official s-Aff-Wild2 images.
	Although these tasks are naturally related through facial behavior, our validation experiments show that they benefit from different visual features, temporal processing strategies, fusion mechanisms, and calibration procedures.
	In this paper, we study task-adaptive feature fusion for ABAW11 multi-task affective behavior analysis.
	We first adapt two pretrained visual backbones, DINOv2 ViT-L and DINOv3 ConvNeXt-base, on an external expression-oriented facial image set, and then freeze them to extract complementary frame-level features from the official ABAW11 data.
	On top of these frozen features, we systematically compare frame-level prediction heads, temporal convolutional heads, post-hoc temporal smoothing, LightGBM models, feature concatenation, gated fusion, residual fusion, late logit fusion, threshold calibration, and shared MTL structures.
	The final system selects task-specific fusion and prediction strategies rather than forcing all tasks to share a single architecture.
	On the ABAW11 validation set, the selected system achieves 0.4222 EXPR macro-F1, 0.5402 AU macro-F1, and 0.6717 VA mean CCC, resulting in an overall validation score of 1.6341.
	The results suggest that task-adaptive fusion of frozen visual features is a simple and effective strategy for ABAW-style multi-task affective behavior analysis.

\end{abstract}
\section{Introduction}
\label{sec:introduction}

Affective behavior analysis in the wild is important for human--computer interaction, intelligent education, mental health assessment, and social robotics.
However, reliable facial affect recognition in unconstrained settings remains challenging because real-world images exhibit substantial variations in identity, pose, illumination, occlusion, and motion blur, together with annotation ambiguity.
The Affective Behavior Analysis in-the-Wild (ABAW) competitions, built on the Aff-Wild and Aff-Wild2 datasets~\cite{kollias2019affwild,kollias2018affwild2,abaw11website}, provide large-scale benchmarks for evaluating affect recognition under such conditions.
The ABAW11 Multi-Task Learning (MTL) Challenge~\cite{abaw11website} considers three tasks on the official s-Aff-Wild2 images: valence--arousal estimation (VA), categorical expression recognition (EXPR), and facial action unit detection (AU).

Although these tasks share the same visual input, they involve different label structures and prediction objectives.
VA is a continuous regression problem, EXPR is a single-label classification problem, and AU detection is a multi-label classification problem.
Our validation experiments further show that the three tasks favor different feature representations, temporal strategies, fusion mechanisms, and calibration procedures.
This motivates us to investigate task-adaptive feature fusion instead of imposing a single downstream architecture on all tasks.

We first fine-tune a DINOv2 ViT-L backbone~\cite{dosovitskiy2021vit,oquab2024dinov2} and a DINOv3 ConvNeXt-base backbone~\cite{simeoni2025dinov3,liu2022convnext} on an external expression-oriented facial image set.
The adapted backbones are then frozen and used to extract complementary frame-level features from the official ABAW11 images.
On top of these features, we compare neural and LightGBM~\cite{ke2017lightgbm} prediction heads, learned and post-hoc temporal processing, feature- and logit-level fusion, threshold calibration, and shared MTL structures.
The final system selects the highest-scoring validation configuration separately for each task.

Our contributions are threefold.
First, we present a task-adaptive fusion framework that allows VA, EXPR, and AU to use different feature sources, temporal strategies, fusion mechanisms, and calibration procedures.
Second, we conduct a systematic comparison of neural and tree-based prediction heads, temporal processing methods, feature- and prediction-level fusion strategies, and shared MTL structures under a common validation protocol.
Finally, the resulting task-adaptive configuration achieves an overall score of 1.6341 on the official ABAW11 validation set, outperforming the shared MTL baselines evaluated under the same protocol.
\section{Related Work}
\label{sec:related}

\subsection{Affective Behavior Analysis in the Wild}

Aff-Wild~\cite{kollias2019affwild} and Aff-Wild2~\cite{kollias2018affwild2} are large-scale in-the-wild datasets for facial affect analysis.
They cover tasks such as valence--arousal estimation, expression recognition, and action unit detection, and have served as the basis of the ABAW benchmark series.
The 7th ABAW Competition introduced an MTL challenge that jointly considered VA, EXPR, and AU on s-Aff-Wild2~\cite{kollias2024abaw7}.
ABAW11 continues this multi-task setting under the official s-Aff-Wild2 protocol~\cite{abaw11website}.
Unlike single-task systems, ABAW MTL must accommodate continuous regression, single-label classification, and multi-label classification over the same visual input.

\subsection{External Data and Visual Representations}

External facial affect datasets are widely used to improve the transfer performance of affect recognition models.
AffectNet~\cite{mollahosseini2017affectnet} provides expression and valence--arousal annotations, while RAF-DB~\cite{li2017rafdb} contains real-world facial images with expression labels.
FER2013~\cite{goodfellow2013fer2013}, FER+~\cite{barsoum2016ferplus}, and CK+~\cite{lucey2010ckplus} are also commonly used for facial expression recognition.
In this work, an image set constructed primarily from AffectNet and RAF-DB is used only for backbone adaptation, while all downstream models are trained and evaluated using the official ABAW11 annotations.

Recent self-supervised models provide transferable visual representations.
DINOv2~\cite{oquab2024dinov2} and DINOv3~\cite{simeoni2025dinov3} learn general-purpose features from large-scale image collections.
Vision Transformers capture interactions among image tokens~\cite{dosovitskiy2021vit}, whereas ConvNeXt retains convolutional inductive biases such as locality~\cite{liu2022convnext}.
We adapt one backbone from each family to obtain complementary frozen features for the ABAW11 tasks.

\subsection{Fusion, Temporal Modeling, and Multi-Task Learning}

Affective behavior recognition systems commonly combine visual feature extraction, temporal modeling, feature fusion, and task-specific prediction~\cite{poria2017affective}.
Existing ABAW methods have used strong visual encoders, temporal processing, prediction smoothing, and task-specific heads~\cite{savchenko2023emotieffnet,zhang2023abaw5fusion}.
Temporal convolutional networks~\cite{bai2018tcn} and inference-time smoothing can incorporate short-term context or stabilize frame-level predictions.

When multiple feature streams are available, gated fusion~\cite{arevalo2017gmu}, tensor fusion~\cite{zadeh2017tfn}, and transformer-based fusion~\cite{tsai2019mult} can model interactions among heterogeneous representations.
Classical multi-task learning instead shares representations across related tasks while retaining task-specific output heads~\cite{caruana1997multitask}.
Rather than assuming that one shared structure is optimal for all outputs, we compare task-adaptive fusion, temporal processing, calibration, and shared MTL baselines within a common frozen-feature framework.

\section{Method}
\label{sec:method}

\subsection{Overview}
\label{sec:method_overview}

Given a cropped-aligned facial frame $\mathbf{x}_t$, ABAW11 requires a valence--arousal vector $\mathbf{y}^{\mathrm{va}}_t\in[-1,1]^2$, an expression label $y^{\mathrm{expr}}_t\in\{1,\ldots,8\}$, and an action-unit vector $\mathbf{y}^{\mathrm{au}}_t\in\{0,1\}^{12}$.
Since the annotations contain task-specific missing labels, a sample is included in training and evaluation for a task only when it is marked as valid for that task.

Our system reuses two frozen expression-adapted representations and selects the downstream expert separately for each task.
For every frame, we extract
\begin{equation}
    \mathbf{z}^{\mathrm{vit}}_t=f_{\mathrm{vit}}(\mathbf{x}_t),
    \qquad
    \mathbf{z}^{\mathrm{cnn}}_t=f_{\mathrm{cnn}}(\mathbf{x}_t),
\end{equation}
where both features are 1024-dimensional.
The expert pool contains frame-level neural heads, temporal models, LightGBM predictors, feature- and prediction-level fusion modules, and shared MTL baselines.

\subsection{Frozen Visual Representations}
\label{sec:frozen_representations}

Before ABAW11 training, we adapt a DINOv2 ViT-L/14~\cite{dosovitskiy2021vit,oquab2024dinov2} and a DINOv3 ConvNeXt-base~\cite{simeoni2025dinov3,liu2022convnext} on an external eight-class expression set constructed from AffectNet~\cite{mollahosseini2017affectnet} and RAF-DB~\cite{li2017rafdb}.
The adaptation data contain no Aff-Wild2 images.
We then discard the expression classifiers, freeze both backbones, and store their frame features offline.
This makes all downstream comparisons use identical visual inputs while avoiding repeated backbone evaluation.
The adaptation and extraction settings are provided in Appendix~\ref{sec:appendix_backbones}.

\subsection{Task-Specific and Temporal Experts}
\label{sec:task_experts}

For each feature stream, we compare linear and multi-layer frame heads with task-dependent output layers.
VA uses a $\tanh$-bounded regressor optimized by MSE and CCC loss; EXPR uses class-weighted cross-entropy with label smoothing; and AU uses weighted binary cross-entropy followed by per-AU threshold calibration.
As a non-neural baseline, we train LightGBM~\cite{ke2017lightgbm} predictors on the frozen ViT-L features.
We also provide each frame with the mean feature of its local temporal neighborhood and its deviation from this mean, allowing the frame-wise model to capture simple contextual changes.

Temporal neural experts project the frozen features to a shared hidden dimension and apply residual temporal convolutional blocks~\cite{bai2018tcn} over overlapping within-video windows.
As a parameter-free alternative, mean or median smoothing is applied to frame predictions without crossing video boundaries.
AU logits are smoothed before sigmoid activation, whereas VA predictions are smoothed directly.
Exact head structures, losses, sequence settings, smoothing windows, and LightGBM parameters are listed in Appendices~\ref{sec:appendix_heads} and~\ref{sec:appendix_temporal}.

\subsection{Task-Adaptive Feature Fusion}
\label{sec:feature_fusion}

We compare three feature-level combinations of the ViT-L and ConvNeXt representations.
Direct concatenation uses $\mathbf{z}^{\mathrm{cat}}_t=[\mathbf{z}^{\mathrm{vit}}_t;\mathbf{z}^{\mathrm{cnn}}_t]$.
Gated fusion learns sample-dependent weights after independently normalizing and projecting the two features, while residual fusion treats the ViT-L representation as the main branch and introduces the ConvNeXt feature through a learned gate.
Each fused representation can be followed by either a frame-level or temporal head.

We also consider late fusion between independently trained experts:
\begin{equation}
    \hat{\mathbf{p}}_t
    =(1-\alpha)\hat{\mathbf{p}}^{\mathrm{vit}}_t
    +\alpha\hat{\mathbf{p}}^{\mathrm{cnn}}_t,
\end{equation}
where $\alpha$ is selected on the validation set.
Fusion operates on logits for EXPR and AU and on continuous predictions for VA; smoothing and AU threshold calibration are applied afterwards when used.
The complete gated and residual formulations are deferred to Appendix~\ref{sec:appendix_fusion}.

\subsection{Shared Multi-Task Baselines}
\label{sec:shared_mtl}

For comparison with task-adaptive selection, we jointly train shared MTL models~\cite{caruana1997multitask} using the same frozen feature families and validity masks.
The evaluated variants comprise a shared MLP with task-specific heads, a task-token model that exchanges information through a lightweight Transformer, and residual task-relation modules with learned gates.
All models optimize the sum of the VA, EXPR, and AU objectives.
Their architectural details are given in Appendix~\ref{sec:appendix_mtl}.

\begin{table}[!b]
	\centering
	\small
	\caption{Frame-level head ablation. EXPR/AU/VA are evaluated by macro-F1, tuned macro-F1, and mean CCC.}
	\label{tab:frame_head_ablation}
	\setlength{\tabcolsep}{4pt}
	\renewcommand{\arraystretch}{1.05}
	\begin{tabular}{llccc}
		\toprule
		Feature & Task & Linear & MLP & BigMLP \\
		\midrule
		ViT-L & EXPR & 0.3534 & 0.3332 & \textbf{0.3651} \\
		ViT-L & AU & 0.5072 & 0.5200 & \textbf{0.5252} \\
		ViT-L & VA & 0.4317 & 0.4375 & \textbf{0.4437} \\
		\midrule
		ConvNeXt & EXPR & 0.3051 & 0.3065 & \textbf{0.3144} \\
		ConvNeXt & AU & 0.4988 & \textbf{0.5057} & 0.5053 \\
		ConvNeXt & VA & 0.4366 & \textbf{0.4603} & 0.4598 \\
		\bottomrule
	\end{tabular}
\end{table}

\section{Experiments and Results}
\label{sec:experiments}

\subsection{Experimental Setting}

We evaluate our method on the official ABAW11 validation set.
All models are trained on the official training annotations and evaluated on the official validation annotations.
For each task, samples with missing labels are ignored during training and evaluation.
EXPR is evaluated by macro-F1 over eight expression classes, AU is evaluated by macro-F1 over 12 action units, and VA is evaluated by the mean CCC of valence and arousal.

The official ABAW11 MTL validation score is computed as:
\begin{equation}
	S =
	\frac{\mathrm{CCC}_{\mathrm{valence}}+\mathrm{CCC}_{\mathrm{arousal}}}{2}
	+
	\frac{F1_{\mathrm{EXPR}}}{8}
	+
	\frac{F1_{\mathrm{AU}}}{12}.
\end{equation}
Here, $F1_{\mathrm{EXPR}}$ denotes the sum of the F1 scores over the eight expression classes, so $F1_{\mathrm{EXPR}}/8$ is equivalent to EXPR macro-F1.
Similarly, $F1_{\mathrm{AU}}$ denotes the sum of the F1 scores over the 12 action units, and $F1_{\mathrm{AU}}/12$ is equivalent to AU macro-F1.
Therefore, all tables in this paper report EXPR macro-F1, AU macro-F1, and mean VA CCC.

For VA estimation, CCC is computed as:
\begin{equation}
	\mathrm{CCC}(x,y) =
	\frac{2\rho\sigma_x\sigma_y}
	{\sigma_x^2+\sigma_y^2+(\mu_x-\mu_y)^2},
\end{equation}
where $x$ and $y$ denote the prediction and ground truth, $\rho$ is their Pearson correlation coefficient, $\mu_x$ and $\mu_y$ are their means, and $\sigma_x^2$ and $\sigma_y^2$ are their variances.

\subsection{Frame-level Head Analysis}

We first compare three frame-level heads: Linear, MLP, and BigMLP.
As shown in Table~\ref{tab:frame_head_ablation}, the preferred head varies across feature sources and tasks.
BigMLP performs best for all ViT-L tasks, while ConvNeXt-base favors BigMLP for EXPR and MLP for AU and VA.

\subsection{Single-feature Expert Analysis}

We then compare single-feature experts using ViT-L features, ConvNeXt-base features, and LightGBM models trained on ViT-L features.
Table~\ref{tab:single_feature_task_ablation} reports frame-level prediction, post-hoc smoothing, and temporal modeling.
For neural frame-level results, we use the best head from Table~\ref{tab:frame_head_ablation}.
ViT-L is stronger for EXPR and AU, while VA benefits clearly from temporal modeling on both deep feature streams.
LightGBM is competitive for AU, indicating that tree-based models can still exploit frozen facial representations.

\begin{table}[t]
	\centering
	\small
	\caption{Single-feature ablation. EXPR/AU/VA are evaluated by macro-F1, tuned macro-F1, and mean CCC.}
	\label{tab:single_feature_task_ablation}
	\setlength{\tabcolsep}{8pt}
	\renewcommand{\arraystretch}{1.08}
	\begin{tabular}{llccc}
		\toprule
		Task & Feature / Model & Frame & Smooth & Temporal \\
		\midrule
		\multirow{3}{*}{EXPR}
		& ViT-L & 0.3651 & 0.3961 & \textbf{0.4211} \\
		& ConvNeXt-base & 0.3144 & 0.3217 & 0.3670 \\
		& LightGBM (ViT-L feat.) & 0.3275 & 0.3677 & 0.3055 \\
		\midrule
		\multirow{3}{*}{AU}
		& ViT-L & 0.5252 & 0.5314 & \textbf{0.5330} \\
		& ConvNeXt-base & 0.5057 & 0.5132 & 0.5170 \\
		& LightGBM (ViT-L feat.) & 0.5162 & 0.5249 & 0.5249 \\
		\midrule
		\multirow{3}{*}{VA}
		& ViT-L & 0.4437 & 0.4781 & \textbf{0.6367} \\
		& ConvNeXt-base & 0.4603 & 0.5222 & 0.6236 \\
		& LightGBM (ViT-L feat.) & 0.3673 & 0.3868 & 0.4682 \\
		\bottomrule
	\end{tabular}
\end{table}

\begin{table}[t]
	\centering
	\small
	\caption{Dual-feature fusion ablation. EXPR/AU/VA are evaluated by macro-F1, tuned macro-F1, and mean CCC.}
	\label{tab:dual_feature_fusion_ablation}
	\setlength{\tabcolsep}{6pt}
	\renewcommand{\arraystretch}{1.05}
	\begin{tabular}{lccc}
		\toprule
		Method & EXPR & AU & VA \\
		\midrule
		Best ViT-L single expert & 0.4211 & 0.5330 & 0.6367 \\
		Best ConvNeXt-base single expert & 0.3670 & 0.5170 & 0.6236 \\
		\midrule
		Concatenation, frame-level & 0.3513 & 0.5275 & 0.4630 \\
		Concatenation, smoothed & \textbf{0.4222} & 0.5356 & 0.6574 \\
		Concatenation, temporal & 0.3837 & 0.5344 & 0.5081 \\
		\midrule
		Gated fusion, frame-level & 0.3086 & 0.5134 & 0.4351 \\
		Gated fusion, smoothed & 0.4213 & 0.5361 & 0.6574 \\
		Gated fusion, temporal & 0.3437 & 0.5202 & 0.4848 \\
		\midrule
		Residual fusion, frame-level & 0.3091 & 0.5138 & 0.4427 \\
		Residual fusion, smoothed & 0.3989 & 0.5386 & \textbf{0.6717} \\
		Residual fusion, temporal & 0.3334 & 0.5225 & 0.4902 \\
		\midrule
		Late logit fusion, smoothed & 0.4086 & \textbf{0.5402} & 0.6588 \\
		\bottomrule
	\end{tabular}
\end{table}

\subsection{Dual-feature Fusion Analysis}

We further combine ViT-L and ConvNeXt-base features using concatenation, gated fusion, residual fusion, and late logit fusion.
As shown in Table~\ref{tab:dual_feature_fusion_ablation}, the best fusion strategy differs across tasks: EXPR prefers concatenation with smoothing, AU obtains the best result from late logit fusion, and VA benefits most from residual fusion with smoothing.

\subsection{Shared MTL Structure Analysis}

We compare several shared MTL structures using the same frozen feature representations.
Among these baselines, the task-token model achieves the highest overall score of 1.3598.
However, the results suggest that a single shared prediction structure cannot fully accommodate the heterogeneous requirements of VA, EXPR, and AU.

\subsection{Validation-Selected Task-Adaptive System}
\label{sec:validation_selected_system}

Based on the official validation experiments, we construct a validation-selected system using the best-performing configuration for each task.
Specifically, EXPR uses concatenated ViT-L and ConvNeXt-base features with smoothing, AU uses late logit fusion with threshold calibration, and VA uses residual feature fusion with smoothing.
As shown in Table~\ref{tab:shared_mtl_ablation}, these configurations achieve 0.4222, 0.5402, and 0.6717 for EXPR, AU, and VA, respectively, yielding an overall validation score of 1.6341.
This result represents the best configuration on the official validation split, while the stability of the selected experts under different video partitions is examined in Section~\ref{sec:five_fold}.

\begin{table}[t]
	\centering
	\small
	\caption{Comparison of shared MTL baselines and the validation-selected task-adaptive system. Overall is the sum of EXPR macro-F1, AU tuned macro-F1, and mean VA CCC. Bold values indicate the best result within each category.}
	\label{tab:shared_mtl_ablation}
	\setlength{\tabcolsep}{5.5pt}
	\renewcommand{\arraystretch}{1.04}
	\begin{tabular}{lcccc}
		\toprule
		Method & EXPR & AU & VA & Overall \\
		\midrule
		ViT-L MTL
		& 0.3345 & 0.5171 & 0.4787 & 1.3303 \\
		
		ConvNeXt MTL
		& 0.2822 & 0.4997 & 0.4918 & 1.2737 \\
		
		Concat MTL
		& 0.3429 & 0.5165 & 0.4918 & 1.3512 \\
		
		Gated MTL
		& 0.3448 & 0.5146 & 0.4881 & 1.3475 \\
		
		Residual MTL
		& 0.3114 & 0.5112 & 0.4389 & 1.2616 \\
		
		Task-token MTL (best shared)
		& \textbf{0.3460}
		& \textbf{0.5176}
		& \textbf{0.4962}
		& \textbf{1.3598} \\
		
		Task-token + dropout
		& 0.3111 & 0.5165 & 0.4501 & 1.2777 \\
		
		Task-token + residual
		& 0.2916 & 0.5140 & 0.4165 & 1.2221 \\
		
		Task-token + gated
		& 0.3113 & 0.5154 & 0.4525 & 1.2792 \\
		
		\midrule
		\textbf{Task-adaptive fusion}
		& \textbf{0.4222}
		& \textbf{0.5402}
		& \textbf{0.6717}
		& \textbf{1.6341} \\
		\bottomrule
	\end{tabular}
\end{table}

\subsection{Cross-Fold Expert Stability}
\label{sec:five_fold}

We further conduct video-level five-fold cross-validation, ensuring that frames from the same video never appear in both the training and validation subsets.
During this more detailed evaluation, we find that the ViT-L temporal convolutional expert performs consistently well for VA and outperforms residual fusion in every fold.
As shown in Table~\ref{tab:five_fold_results}, the temporal expert achieves $0.6411 \pm 0.0309$, compared with $0.5124 \pm 0.0378$ for residual fusion.
Although residual fusion obtains a higher score on the official validation split, the five-fold results indicate that explicit temporal modeling is more stable across different video partitions.

Based on this observation, our final system retains concatenated features for EXPR and late fusion for AU, while replacing the residual VA expert with the ViT-L temporal convolutional expert.
The resulting task-adaptive configuration achieves an overall five-fold score of $1.6166 \pm 0.0394$.

\begin{table}[t]
	\centering
	\small
	\caption{Video-level five-fold results. VA-Res. denotes residual fusion with smoothing. Overall is computed using VA-TCN.}
	\label{tab:five_fold_results}
	\renewcommand{\arraystretch}{0.90}
	\begin{tabular*}{\linewidth}
		{@{\extracolsep{\fill}}lccccc@{}}
		\toprule[1pt]
		Fold & EXPR & AU & VA-Res. & VA-TCN & Overall \\
		\midrule
		1 & 0.5149 & 0.5648 & 0.5360 & \textbf{0.6066} & 1.6863 \\
		2 & 0.4189 & 0.5639 & 0.5086 & \textbf{0.6260} & 1.6089 \\
		3 & 0.3613 & 0.5661 & 0.5584 & \textbf{0.6982} & 1.6256 \\
		4 & 0.4070 & 0.5197 & 0.5135 & \textbf{0.6432} & 1.5699 \\
		5 & 0.4109 & 0.5502 & 0.4456 & \textbf{0.6312} & 1.5924 \\
		\midrule
		Mean
		& 0.4226
		& 0.5530
		& 0.5124
		& \textbf{0.6411}
		& 1.6166 \\
		SD
		& 0.0503
		& 0.0176
		& 0.0378
		& \textbf{0.0309}
		& 0.0394 \\
		\bottomrule[1pt]
	\end{tabular*}
\end{table}
\section{Discussion and Conclusion}
\label{sec:discussion_conclusion}

Our results show that ABAW11 tasks favor different downstream configurations.
EXPR benefits from concatenated ViT-L and ConvNeXt-base features, while AU favors late fusion and threshold calibration.
For VA, residual fusion achieves the highest score on the official validation split, whereas the temporal convolutional expert is more stable across video-level folds.
The resulting cross-validated configuration obtains $1.6166 \pm 0.0394$ and remains stronger than the shared MTL baselines, supporting task-specific expert selection.

Although the three tasks operate on the same facial frames, their labels depend on different information.
VA is closely related to temporal affective evolution, EXPR emphasizes holistic categorical patterns, and AU focuses on localized facial activations.
These differences may explain their preferences for different representations, temporal strategies, and calibration procedures.

The current system relies on frozen backbones and validation-based expert selection.
Future work could investigate these task-specific preferences and develop automatic expert weighting or joint backbone adaptation.
Since the official competition results are not yet available, all conclusions are based on validation experiments.

\FloatBarrier
\bibliographystyle{splncs04}
\bibliography{main}

\clearpage
\appendix
% This file is enabled by \includeappendixtrue in main.tex.
\appendix
\renewcommand*{\theHsection}{appendix.\Alph{section}}
\renewcommand*{\theHsubsection}{appendix.\Alph{section}.\arabic{subsection}}

\section{Implementation Details}
\label{sec:appendix_implementation}

\subsection{Overall Pipeline and Feature Extraction}
\label{sec:appendix_pipeline}

\begin{figure}[!b]
	\centering
	\includegraphics[width=0.92\linewidth]{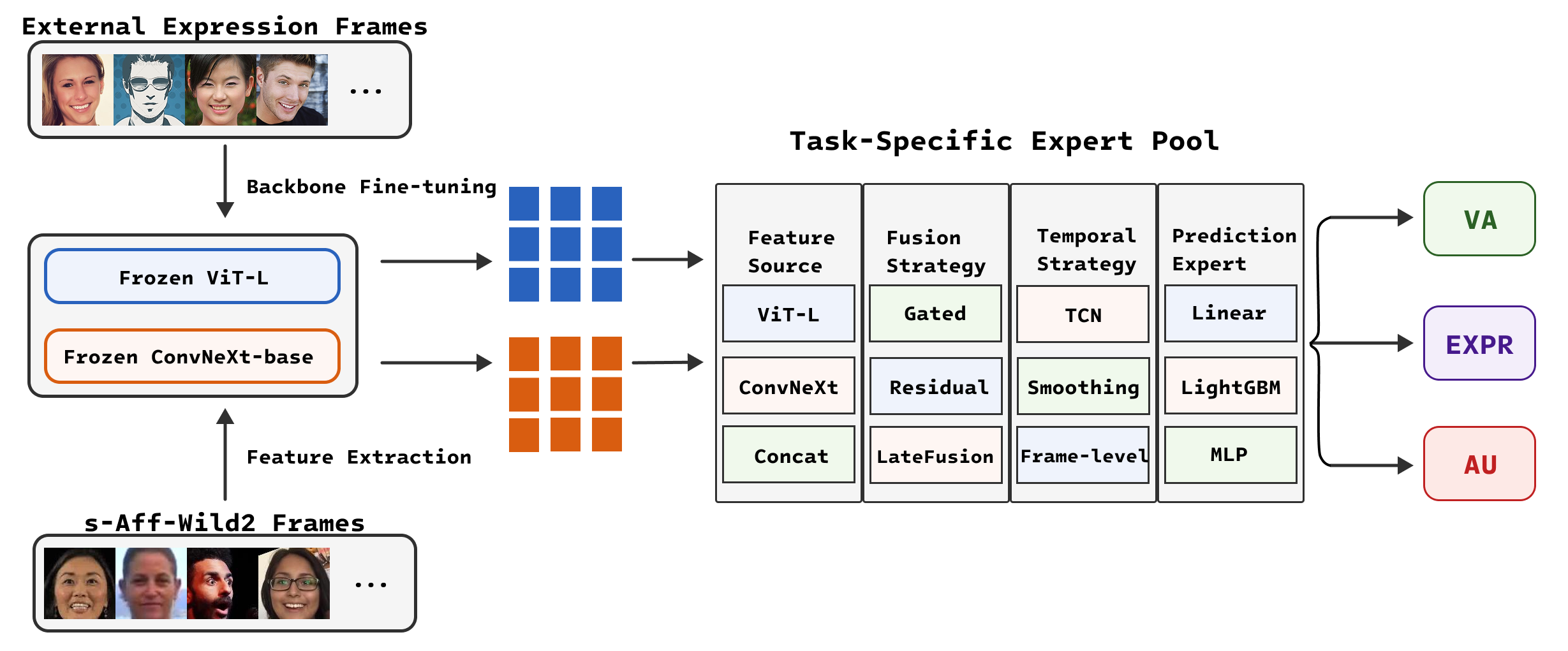}
	\caption{Overview of the proposed task-specific expert framework.
		The two visual backbones are first adapted using external expression frames and are subsequently frozen for feature extraction on s-Aff-Wild2.
		The extracted representations are supplied to a pool of candidate feature sources, fusion strategies, temporal strategies, and prediction experts.
		The components inside the expert pool represent alternative task-specific choices rather than a single sequential architecture.}
	\label{fig:main_structure}
\end{figure}

Figure~\ref{fig:main_structure} presents the overall workflow of our system.
The pipeline consists of three main stages: external-data backbone adaptation, frozen feature extraction on s-Aff-Wild2, and task-specific expert construction.
Instead of training a single end-to-end architecture for all three ABAW11 tasks, we first establish two complementary visual representations and then construct separate downstream configurations for VA, EXPR, and AU.

\paragraph{Backbone adaptation.}
We use a DINOv2 ViT-L and a ConvNeXt-base as the two visual backbones.
Before processing s-Aff-Wild2, both models are adapted using external facial-expression frames.
For ViT-L, the patch embedding and the first eight Transformer blocks are kept fixed during adaptation.
The remaining representation is optimized with a four-expert mixture-of-experts head, where each expert contains three residual MLP blocks.
The mixture-of-experts prediction head is used only during external-data adaptation and is discarded afterward.
The normalized 1024-dimensional class-token representation is retained as the ViT-L feature.

ConvNeXt-base is adapted separately using a conventional expression-classification objective.
After adaptation, its classification head is also removed, and the 1024-dimensional globally pooled representation is retained.
The two adaptation procedures are independent; no ABAW11 validation or test annotations are used at this stage.
This produces two visual encoders with different inductive biases: ViT-L provides a global attention-based representation, whereas ConvNeXt-base provides a convolutional representation with stronger spatial locality.

\subsection{Backbone Adaptation and Feature Extraction}
\label{sec:appendix_backbones}

The external expression set used for backbone adaptation is constructed from AffectNet~\cite{mollahosseini2017affectnet} and RAF-DB~\cite{li2017rafdb} and reorganized into neutral, anger, disgust, fear, happiness, sadness, surprise, and other.
This stage contains no Aff-Wild2 images.

For the transformer branch, we initialize DINOv2 ViT-L/14~\cite{dosovitskiy2021vit,oquab2024dinov2} from pretrained weights.
The patch embedding layer and first eight transformer blocks are frozen, while the remaining layers are adapted through expression classification.
A four-expert gated mixture-of-experts classifier is attached during adaptation; each expert contains three residual MLP blocks, and a learned router produces sample-dependent softmax weights.
Training uses class-balanced cross-entropy, label smoothing, MixUp, layer-wise learning-rate decay, and facial-image augmentation.
After adaptation, the classifier is discarded and the normalized class token is retained as a 1024-dimensional representation.

For the convolutional branch, we initialize DINOv3 ConvNeXt-base~\cite{simeoni2025dinov3,liu2022convnext} and fine-tune the backbone together with a linear expression classifier.
It uses the same eight classes, class-balanced supervision, label smoothing, MixUp, and image augmentation, with a separately configured optimization schedule.
The classifier is then removed and the 1024-dimensional global representation is retained.

For ABAW11 extraction, the cleaned cropped-aligned frames are resized to $256\times256$, center-cropped to $224\times224$, and normalized with ImageNet statistics.
Both backbones are evaluated without gradient updates, and the resulting features are stored offline and matched to annotations by relative frame path.

\subsection{Prediction Heads, Objectives, and LightGBM}
\label{sec:appendix_heads}

The linear head consists of layer normalization, dropout, and a single output projection.
The MLP head uses one or two GELU-activated hidden layers depending on the task, while BigMLP uses two wider hidden layers with stronger input and hidden dropout.
The output dimensions are two, eight, and twelve for VA, EXPR, and AU, respectively.
VA predictions are bounded by $\tanh$, whereas EXPR and AU heads output logits.

The VA objective combines mean squared error and concordance correlation coefficient loss:
\begin{equation}
	\mathcal{L}_{\mathrm{va}}
	=
	\lambda_{\mathrm{mse}}
	\operatorname{MSE}
	\left(
	\hat{\mathbf{y}}^{\mathrm{va}},
	\mathbf{y}^{\mathrm{va}}
	\right)
	+
	\frac{\lambda_{\mathrm{ccc}}}{2}
	\sum_{d\in\{v,a\}}
	\left(1-\mathrm{CCC}_d\right),
\end{equation}
where $\lambda_{\mathrm{mse}}=0.5$ and $\lambda_{\mathrm{ccc}}=1$.
For EXPR, the base class weight is $N/(8N_c)$, where $N_c$ is the number of samples in class $c$.
Its square root is normalized by the mean class weight and used in cross-entropy with label smoothing.
For AU $j$, the positive weight is the square root of its negative-to-positive sample ratio and is used in binary cross-entropy with logits.

As a non-neural baseline, we also train LightGBM~\cite{ke2017lightgbm} predictors on the frozen ViT-L features.
We use two regressors for valence and arousal, one multiclass classifier for EXPR, and twelve binary classifiers for AU.
Each model uses 300 estimators, a learning rate of 0.03, at most 31 leaves, a subsampling ratio of 0.9, a feature-sampling ratio of 0.8, and a random seed of 42.

To provide the frame-wise LightGBM models with simple temporal context, we additionally compute the local mean feature and the deviation of the current frame from this mean:
\begin{equation}
	\boldsymbol{\mu}_t
	=
	\frac{1}{|\mathcal{N}_t|}
	\sum_{i\in\mathcal{N}_t}\mathbf{z}_i,
	\qquad
	\boldsymbol{\delta}_t
	=
	\mathbf{z}_t-\boldsymbol{\mu}_t,
\end{equation}
where $\mathcal{N}_t$ contains frames from the same video within a radius of 12.
The temporally augmented input is then
\begin{equation}
	\tilde{\mathbf{z}}_t
	=
	[\mathbf{z}_t;
	\boldsymbol{\mu}_t;
	\boldsymbol{\delta}_t].
\end{equation}

\subsection{Temporal Modeling and Post-processing}
\label{sec:appendix_temporal}

Task-valid frames are ordered within each video and divided into overlapping sequences.
EXPR and AU use sequences of 256 frames with a stride of 128, while VA uses sequences of 512 frames with a stride of 256.
Short sequences are zero-padded, and padded positions are excluded from both prediction and loss computation.

Each input feature is first normalized and projected to a 256-dimensional hidden representation.
The resulting sequence is processed by residual temporal convolutional blocks:
\begin{equation}
	\mathbf{H}^{(l+1)}
	=
	\mathbf{H}^{(l)}
	+
	\mathcal{F}_{l}\left(\mathbf{H}^{(l)}\right),
\end{equation}
where $\mathcal{F}_{l}$ contains two one-dimensional convolutions with kernel size five, group normalization, GELU activation, and dropout.
The dilation factor increases exponentially across blocks as $1,2,4,\ldots$.
We use four blocks for EXPR and AU and six blocks for VA.
No Transformer layer is used in the final VA temporal expert.
Predictions of frames appearing in multiple overlapping sequences are averaged.

As a parameter-free temporal alternative, we apply mean or median smoothing independently within each video.
For a smoothing window of size $w$, the prediction at frame $t$ is computed from the neighboring predictions:
\begin{equation}
	\tilde{\mathbf{p}}_t
	=
	\operatorname{Aggregate}
	\left(
	\mathbf{p}_{t-r},\ldots,\mathbf{p}_{t+r}
	\right),
	\qquad
	r=\frac{w-1}{2},
\end{equation}
where $\operatorname{Aggregate}$ denotes the mean or median operator.
Boundary predictions are replicated, and smoothing never crosses video boundaries.
We evaluate window sizes of $3,5,7,9,13,17,$ and $25$; the standard EXPR configuration uses median smoothing with a window size of 25.
AU logits are smoothed before sigmoid activation, whereas VA predictions are smoothed directly.

For AU threshold calibration, the decision threshold of each AU is selected independently by maximizing its validation F1 score.
Thresholds are searched from 0.01 to 0.99 with a step size of 0.01.

\subsection{Trainable Fusion Architectures}
\label{sec:appendix_fusion}

For trainable feature fusion, the frozen ViT-L and ConvNeXt-base features are independently normalized and projected to the same dimensional space:
\begin{equation}
	\mathbf{a}_t
	=
	W_{\mathrm{vit}}
	\mathrm{LN}
	\left(\mathbf{z}^{\mathrm{vit}}_t\right),
	\qquad
	\mathbf{b}_t
	=
	W_{\mathrm{cnn}}
	\mathrm{LN}
	\left(\mathbf{z}^{\mathrm{cnn}}_t\right).
\end{equation}

For gated fusion, a two-dimensional softmax gate first estimates the relative contribution of the two feature streams:
\begin{align}
	\boldsymbol{\omega}_t
	&=
	\mathrm{softmax}
	\left(
	G_{\mathrm{mix}}
	\left(
	[\mathbf{a}_t;\mathbf{b}_t]
	\right)
	\right),\\
	\mathbf{m}_t
	&=
	\omega^{\mathrm{vit}}_t\mathbf{a}_t
	+
	\omega^{\mathrm{cnn}}_t\mathbf{b}_t.
\end{align}
A scalar residual gate then controls how strongly the mixed representation modifies the ViT-L branch:
\begin{align}
	g_t
	&=
	\sigma
	\left(
	G_{\mathrm{res}}
	\left(
	[\mathbf{a}_t;\mathbf{b}_t]
	\right)
	\right),\\
	\mathbf{z}^{\mathrm{gate}}_t
	&=
	\mathrm{LN}
	\left(
	\mathbf{a}_t
	+
	g_t
	\left(
	\mathbf{m}_t-\mathbf{a}_t
	\right)
	\right).
\end{align}

Residual fusion directly adds the projected ConvNeXt-base representation to the ViT-L branch:
\begin{equation}
	\mathbf{z}^{\mathrm{res}}_t
	=
	\mathrm{LN}
	\left(
	\mathbf{a}_t+g_t\mathbf{b}_t
	\right).
\end{equation}
The residual-gate bias is initialized to $-4$, causing the model to begin training close to the ViT-L representation and gradually incorporate complementary ConvNeXt-base information.
Fusion is performed independently at each frame, after which the fused representation is passed to either a frame-level prediction head or a temporal convolutional head.

\subsection{Shared Multi-Task Baselines}
\label{sec:appendix_mtl}

The shared MTL models use ViT-L, ConvNeXt-base, or concatenated features.
For each feature source, the input is standardized using the mean and standard deviation computed from the training set.

The shared-MLP baseline projects each frame feature to a 512-dimensional common representation.
Three task-specific MLP heads then predict VA, EXPR, and AU from the same shared representation.

The task-token model first projects the frame feature to a 512-dimensional representation.
Three task-specific adapters transform this representation, and a learned task embedding is added to each adapted feature.
The resulting VA, EXPR, and AU tokens are processed jointly by a one-layer Transformer encoder with four attention heads:
\begin{equation}
	[\tilde{\mathbf{h}}^{\mathrm{va}}_t;
	\tilde{\mathbf{h}}^{\mathrm{expr}}_t;
	\tilde{\mathbf{h}}^{\mathrm{au}}_t]
	=
	\operatorname{Transformer}
	\left(
	[\mathbf{h}^{\mathrm{va}}_t;
	\mathbf{h}^{\mathrm{expr}}_t;
	\mathbf{h}^{\mathrm{au}}_t]
	\right).
\end{equation}
Separate prediction heads decode the three updated task representations.

The residual-relation model converts each frame feature into 64 latent tokens of dimension 64.
These are feature-space tokens rather than consecutive video frames.
A lightweight depthwise convolutional token mixer produces a shared update, which is added to the original input through task-specific sigmoid gates and learned residual scales:
\begin{equation}
	\mathbf{h}^{k}_t
	=
	\mathbf{x}_t
	+
	\tanh(\alpha_k)
	\mathbf{g}^{k}_t
	\odot
	\Delta\mathbf{x}_t,
	\qquad
	k\in\{\mathrm{va},\mathrm{expr},\mathrm{au}\}.
\end{equation}
Each task representation is finally passed to its corresponding output head.
All shared MTL baselines operate independently at the frame level and do not introduce explicit temporal modeling.

Since not every frame provides valid annotations for all three tasks, each task loss is computed only over its valid samples.
The joint objective is
\begin{equation}
	\mathcal{L}_{\mathrm{MTL}}
	=
	\lambda_{\mathrm{va}}\mathcal{L}_{\mathrm{va}}
	+
	\lambda_{\mathrm{expr}}\mathcal{L}_{\mathrm{expr}}
	+
	\lambda_{\mathrm{au}}\mathcal{L}_{\mathrm{au}},
\end{equation}
where all task weights are set to one.

\end{document}